\documentclass[10pt,twocolumn,letterpaper]{article}

\newif\ifarxiv
\arxivtrue 

\ifarxiv
\newcommand{\IEEEmembership}[1]{}

\usepackage[letterpaper, top=0.6in, bottom=0.5in, inner=0.75in, outer=0.75in, headheight=0.3in, footskip=0.3in]{geometry}
\usepackage{fancyhdr}
\usepackage{times}

\else
    \documentclass[journal]{IEEEtran}
\fi

\usepackage{wacv}
\usepackage{times}
\usepackage{epsfig}
\usepackage{comment}
\usepackage{graphicx, array, multirow}
\usepackage{amsmath}
\usepackage{amssymb}
\usepackage[accsupp]{axessibility} 
\DeclareMathOperator*{\argmax}{argmax} 
\DeclareMathOperator*{\argmin}{argmin} 

\wacvfinalcopy 

\ifwacvfinal
\def\assignedStartPage{9876} 
\fi

\ifwacvfinal
\usepackage[breaklinks=true,bookmarks=false]{hyperref}
\else
\usepackage[pagebackref=true,breaklinks=true,colorlinks,bookmarks=false]{hyperref}
\fi

\ifwacvfinal
\else
\fi

\begin{document}

\title{Uncertainty-Aware Regularization for Image-to-Image Translation}
\author{Anuja Vats, Ivar Farup, Marius Pedersen, Kiran Raja \\
Department of Computer Science, NTNU, Gjøvik, Norway\\}

\maketitle

\begin{abstract}
The importance of quantifying uncertainty in deep networks has become paramount for reliable real-world applications. In this paper, we propose a method to improve uncertainty estimation in medical Image-to-Image (I2I) translation. Our model integrates aleatoric uncertainty and employs Uncertainty-Aware Regularization (UAR) inspired by simple priors to refine uncertainty estimates and enhance reconstruction quality. We show that by leveraging simple priors on parameters, our approach captures more robust uncertainty maps, effectively refining them to indicate precisely where the network encounters difficulties, while being less affected by noise. Our experiments demonstrate that UAR not only improves translation performance, but also provides better uncertainty estimations, particularly in the presence of noise and artifacts. We validate our approach using two medical imaging datasets, showcasing its effectiveness in maintaining high confidence in familiar regions while accurately identifying areas of uncertainty in novel/ambiguous scenarios. 
\end{abstract}

\section{Introduction}
\label{sec:intro}

The significance of quantifying the uncertainty embedded in the learning process of deep neural networks has become increasingly important for identifying the blind spots \cite{NHTSA} and biases \cite{guynn2015google} in models before their application in the real world. Despite the quest for learning from datasets that are larger, more diverse and representative, it has become evident that not all data points will adhere to the assumed distribution, leaving room for potential inaccuracies and biases in model predictions. Model uncertainty as a measure can serve to be a very useful tool for identifying the limitations to our model predictions and accounting for instances where real data may lie outside the learning distribution.
Considering critical application domains such as healthcare, military, criminal justice or automated driving for deep learning models, model performance, albeit high in isolation, is being considered grossly insufficient for their adoption in practice \cite{schneeberger2020european}. The inability to isolate scenarios where a model isn't confident about its decision and the causes underlying that poses a significant barrier to trust and reliability in these safety-critical domains.

This article discusses uncertainty in the context of medical I2I translation. I2I is the problem of transforming an image from one domain into a corresponding image in another domain while maintaining semantic consistency and information preservation during translation. In traditional endoscopy, narrowband imaging (NBI) is used alongside standard imaging for enhanced visualization of abnormalities. However, if a region is not captured with NBI during the procedure, the enhanced information is unavailable post-procedure, limiting its use in retrospective diagnosis. Models that translate standard images to NBI are therefore critical, offering a valuable tool for post-hoc analysis when NBI was not captured. While similar functionality would be highly beneficial in capsule endoscopy, constraints like device size and battery limit real-time acquisition. Virtual chromo-endoscopy (e.g., FICE) can be applied post-capture to enhance abnormalities  \cite{sato2014clinical}, making I2I translation particularly relevant in endoscopic imaging.
Despite advancements, deep learning methods for I2I translation often produce outputs with inherent uncertainties, particularly in ambiguous or unseen scenarios. Moreover, attempts at generalization exacerbates this uncertainty, as variations in datasets or slight shifts in capture modalities can rapidly escalate uncertainty levels. Since, medical image acquisition is often prone to noise and modality-specific artifacts, it is paramount to faithfully quantify and convey model uncertainty to ascertain the extent of generalization achievable. Delineating the model's confidence levels and identifying domain gaps where it struggles, allows to effectively discern where and how to apply the model.

Model uncertainty is broadly composed to two types, the \textit{epistemic} or uncertainty regarding the model parameters and \textit{aleatoric} resulting from noise inherent in the data \cite{der2009aleatory, kendall2017uncertainties}. The epistemic uncertainty assumes a prior distribution over model parameters and often approximated as the variance in predictions from multiple forward passes through the network with different dropout masks applied for example. The aleatoric uncertainty, on the other hand, assumes a distribution on the models outputs and is approximated using Maximum a Posteriori (MAP) estimation \cite{nix1994estimating}. It has been shown that incorporating aleatoric uncertainty during learning can provide useful guidance for learning, especially in high-data regimes \cite{kendall2017uncertainties}. 

In this work, we aim to develop an end-to-end model for I2I translation that incorporates aleatoric uncertainty. Our primary goal is to demonstrate that imposing regularization constraints on the assumed prior distribution can improve estimation of aleatoric uncertainty during the translation process. Additionally, we show that this regularization not only provides more robust uncertainty maps but also improves the overall reconstruction quality (Table \ref{fig:resul}) , affirming that uncertainty-estimation effectively serves as guidance for improved translation \cite{kendall2017uncertainties}. This approach offers an advantage over previous multi-stage architectures \cite{upadhyay2021uncertainty} by eliminating the need for sequential uncertainty estimates between models. Instead, we aim to incorporate a cost-effective regularization term directly into the optimization, facilitating concurrent and mutually beneficial refinement of uncertainty and image translation within a single model.

Our main contributions are (a) a simple and model-agnostic Uncertainty-Aware Regularization (UAR), and (b) a new paired dataset for I2I translation from RGB to FICE in capsule endoscopy. Despite its simplicity, UAR not only improved translation performance (Sections \ref{sec:results} and \ref{sec:ablation_losses}) but allows a more faithful estimation of data-driven uncertainty in the face of commonly encountered noise-corruptions (Section \ref{sec:results}). Finally, UAR shows improved uncertainty prediction in the presence of unforeseen structures/artifacts (Section \ref{sec:artefacts}). 
To understand the effects of various design choices, we conduct ablation experiments in Section \ref{sec:ablation}. 


\section{Related Work}
\label{sec:related_work}

Medical image-to-image translation has seen significant advancements through the use of generative adversarial networks (GANs) and its variants. Typical application in medical I2I include modality translation \cite{armanious2020medgan, yang2020mri}, image synthesis \cite{zhang2018translating}, segmentation \cite{platscher2022image} and super-resolution \cite{gu2020medsrgan}. Modality translation using CycleGAN \cite{zhu2017unpaired} has been particularly influential, enabling unsupervised translation by employing cycle consistency losses to ensure that translated images can be mapped back to the original modality. Similarly, conditional GANs \cite{isola2017image} have allowed generation preconditioned on inputs such as anatomical labels \cite{amirrajab2022label}, modality \cite{dar2019image} or priors useful to generation \cite{amirkolaee2022medical}. Another class of models includes diffusion models \cite{ho2020denoising, kazerouni2023diffusion} that utilize parameterized Markov chains to iteratively refine data, optimizing the lower variational bound on the likelihood function \cite{li2016precomputed, du2023arsdm}. 
In WCE, image translation has been most commonly applied for image super-resolution \cite{almalioglu2020endol2h, turan2021generative}. 
Uncertainty quantification in medical I2I has been relatively less explored. In ~\cite{reinhold2020validating} authors argue the usefulness of uncertainty estimation in MR to CT translation for detecting synthesis failures. They use traditional formulations where epistemic uncertainty is estimated by sampling from a variational distribution using dropout, and the aleatoric component is derived from the variance of the predicted distribution. Authors in \cite{ayhan2018test} and \cite{baltruschat2024uncertainty} utilize variations of test-time augmentation for estimating uncertainty. Ayhan et al. \cite{ayhan2018test} generate augmented examples for each test case to approximate the predictive distribution, whereas Baltruschat et al. leverage predictions from multiple 2D slicing planes instead of augmentations for the same goal. Our work is most closely related to \cite{upadhyay2021uncertainty, upadhyay2021uncertain, upadhyay2021robustness} that model predictive distributions using generalized Gaussian distributions. However, unlike \cite{upadhyay2021uncertainty}, which employs multiple sequential GANs to iteratively reduce aleatoric uncertainty, we introduce a lightweight regularization term that achieves this within a single model. 
As a result, our uncertainty estimates can differentiate between familiar versus newer or significantly larger sources of uncertainty,  overcoming the drawback of previous methods that treat all uncertainty sources equally.


One of the primary challenges in medical I2I translation problems is the inherent ambiguity associated with image capturing mechanisms and its effect on a model's performance. Consider the case of WCE where images are often captured using low-resolution cameras under myriad distortions \cite{yung2020poor, ali2021deep}, requiring significant post-processing before they are suitable for diagnosis. Noise and compression artifacts encountered during transmission further degrade the quality \cite{floor2020error}. The cumulative impact of these factors can manifest subtly as deviations from the anticipated model performance, potentially leading to misdiagnoses. As discussed prior, one approach to mitigating this is to quantify the uncertainty associated with model predictions. Measuring the uncertainty allows detecting unaccounted shifts that can be addressed proactively. Despite relevance, uncertainty quantification and refinement is relatively nascent in I2I translation problems.



\begin{figure*}
   \centering
    \includegraphics[width=16cm,height=5.2cm]{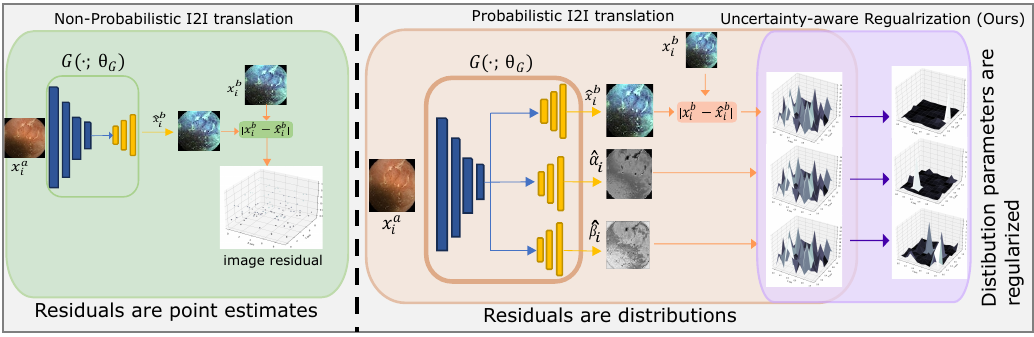}
    \caption{Non-probabilistic image translation (green) optimizes point-estimates for the residuals between the predicted and the target images. On the contrary, the probabilistic approach (orange) models the residuals with a distribution, allowing the variance of errors to change spatially. Our method takes this a step further (purple) by regularizing the predicted variances (or distribution parameters) to achieve more precise uncertainty estimation (The discriminator is omitted from the diagram for simplicity).}
    \label{fig:blockdiag}
\end{figure*}

\section{Methodology}
We introduce both the conventional and probabilistic formulations of paired I2I translation, highlighting their limitations. Subsequently, we present the proposed UAR for improving uncertainty estimation and guidance.

\subsection{I2I Translation Formulation}
Consider a collection of input images from a domain $\mathcal{A}$ denoted as $\mathcal{X}_{\mathcal{A}} := \{x^a_1, x^a_2, ..., x^a_n\}$, and another set of paired images originating from a domain $\mathcal{B}$, expressed as $\mathcal{X}_{\mathcal{B}} := \{x^b_1, x^b_2, ..., x^b_n\}$. The dataset $\mathcal{D}$ comprises pairs $(x^a_i,x^b_i)$ drawn from the respective domains $\mathcal{A}$ and $\mathcal{B}$. The objective is to learn the underlying conditional distribution $\mathcal{P}_{\mathcal{B}|\mathcal{A}}$ facilitating the translation of images from $\mathcal{A} \rightarrow\mathcal{B}$.

As shown in Fig.\ref{fig:blockdiag}, this can typically be achieved by minimizing the point estimate for per-pixel residual at $jk$, $\delta_{jk} = ||\hat{x}^b_{jk} - x^b_{jk}||^2$ between the reconstructed and ground-truth image from domain $\mathcal{B}$. However, in pixel reconstruction tasks, the solution space is often multimodal, meaning that multiple outputs can yield acceptable solutions. Thus, relying solely on point-wise estimation fails to adequately represent the distribution $\mathcal{P}_{\mathcal{B}|\mathcal{A}}$ over the output space, as well as estimate the uncertainty associated with the reconstruction process. The probabilistic remedy for this is to relax the constraint on the residual by modeling it as a distribution instead of a point estimate, the optimal parameters of which are learned from the data, thus allowing an estimation of the uncertainty. As an example, consider a deep learning model  $\mathcal{F}(\mathcal{D};\theta)$ parametrized by $\theta$ to be trained for translating images from domain $\mathcal{A} \rightarrow\mathcal{B}$. While one conceivable distribution for \(\delta\) might be an isotropic standard Gaussian, presuming a fixed variance not only imposes an assumption of independence and identical distribution (i.i.d.) on the residuals, which can be easily compromised by slightly out-of-distribution samples \cite{kendall2017uncertainties, upadhyay2021uncertainty}, but also eliminates the ability to model heteroscedasticity in predictions. Alternatively, the distribution over $\delta$ can be heteroscedastic Gaussian \cite{upadhyay2021uncertainty} with zero mean and spatially varying-learnable standard deviation $\sigma_{jk}$ as in Eq. \ref{eq:1},
\begin{equation}
\label{eq:1}
    \hat{x}_{jk} = x_{jk} + \delta_{jk}, \quad \delta_{jk} \sim \mathcal{N}(0, \sigma^2_{jk}); \quad \hat{x}_{jk} \sim \mathcal{N}(x_{jk}, \sigma^2_{jk})
\end{equation}
The parameters of the network $\mathcal{F}(\mathcal{D};\theta)$ can be optimized by maximizing the likelihood given by:
\begin{equation}
\label{eq:2}
\begin{matrix}
    \mathcal{L}(\mathcal{D};\theta) := {\displaystyle \prod_{i=1}^{n} \mathcal{P}_{\mathcal{B}|\mathcal{A}}(x^b_{i}; \{\hat{x}^b_{i}, \hat{\sigma_i}\})}\\
    \theta^{*} :=\argmax\limits_{\theta}\mathcal{L}(\mathcal{D}; \theta) \\
     = \argmax\limits_{\theta} \displaystyle \prod_{i=1}^{n} \mathcal{P}_{\mathcal{B}|\mathcal{A}}(x^b_{i}; \{\hat{x}^b_{i}, \hat{\sigma_i}\}) \\
    \theta^{*}  = \argmax\limits_{\theta} \displaystyle \prod_{i=1}^{n} \frac{1}{\sqrt{2\pi\hat{\sigma}_i^2}} e^{-\frac{|\hat{x}_i^b - x_i^b|^2}{2\hat{\sigma}^2_i}} \\
\end{matrix}
\end{equation}
where we omit spatial indices $jk$ for simplicity. The negative log likelihood is,
\begin{equation}
\label{eq:3}
    \theta^{*} = \argmin\limits_{\theta} \displaystyle \sum_{i=1}^{n} \left\{\frac{|\hat{x}_i^b - x_i^b|^2}{2\hat{\sigma}^2_i} + \frac{\log(\hat{\sigma}_i^2)}{2}\right\}.\\
\end{equation}
\begin{figure*}
   \centering
    \includegraphics[width=16cm,height=5cm]{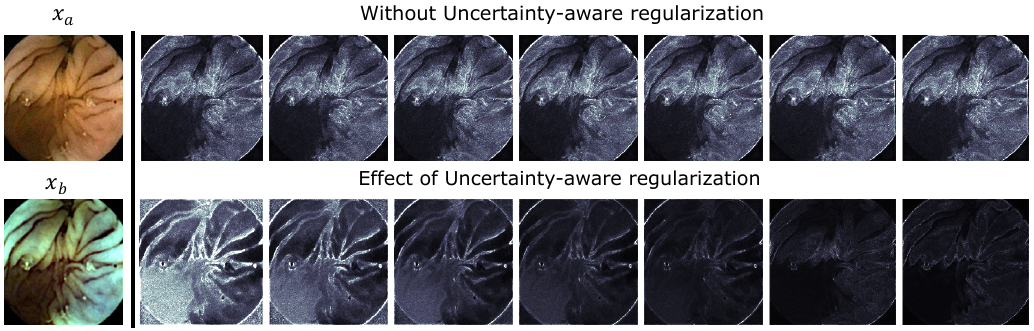}
    \caption{The figure shows the shape $\beta$ parameter and predicted aleatoric uncertainty of an image at different epochs during the training. Without regularization (first row), the variances in the predictions (uncertainty) remain relatively the same throughout training. In contrast, with regularization (second row), the predicted uncertainty gets progressively less noisy and more semantically refined over the course of the training.}
    \label{fig:regularization_effect}
\end{figure*}
Assuming that the residuals follow a normal distribution simplifies uncertainty estimation, as the per-pixel variance $\sigma^2_i$ itself is the aleatoric uncertainty in prediction. This formulation for modeling aleatoric uncertainty can be improved by assuming a more lenient Generalized Normal Distribution (GND) with zero mean over the residuals \cite{upadhyay2021uncertainty, upadhyay2021robustness}. The parameters governing the shape $(\beta)$ and scale $(\alpha)$ of the predicted distribution not only accommodate the heteroscedastic variations in residuals, but also enable heavier-tails, which are beneficial for handling outliers. 
\begin{equation}
\label{eq:4}
     \delta_{jk} \sim GND(\delta; 0,\alpha_{jk},\beta_{jk})\\      
\end{equation}
As before, the likelihood can be written as:
\begin{equation}
\begin{matrix}
\label{eq:5}
 \mathcal{L}(\mathcal{D};\theta) := {\displaystyle \prod_{i=1}^{n} \mathcal{P}_{\mathcal{B}|\mathcal{A}}(x^b_{i}; \{\hat{x}^b_{i}, \hat{\alpha_i}, \hat{\beta_i}\})}\\
 \theta^{*} :=\argmax\limits_{\theta}\mathcal{L}(\mathcal{D}; \theta) \\
  \theta^{*} = \arg\max\limits_{\theta} \displaystyle \prod_{i=1}^{n} \frac{\hat{\beta}_i}{2\hat{\alpha}_i\Gamma(\frac{1}{\beta}_i)} e^{-\left(\frac{|\hat{x}_i^b - x_i^b|}{\hat{\alpha}_i}\right)^{\hat{\beta}_i}}
    \end{matrix}
\end{equation}
Therefore, the negative likelihood is,
\begin{equation}
\label{eq:6}
     \theta^{*} = \argmin\limits_{\theta} \displaystyle \sum_{i=1}^{n} \left\{\left( \frac{|\hat{x}_i^b - x_i^b|}{\hat{\alpha}_i}\right)^{\hat{\beta}_i} - \log \frac{\hat{\beta}_i}{\hat{\alpha}_i} + \log\ \Gamma(\frac{1}{\hat{\beta}}_i)\right\}
\end{equation}
We refer to this loss as the negative likelihood loss $L_{nll}$, in next sections. The aleatoric uncertainty for $\hat{x}_i^b$ can be written down as the variance of this distribution, given by $\frac{\hat{\alpha}_i^2\ \Gamma(3/\hat{\beta}_i)}{\Gamma(1/\hat{\beta}_i)}$. The generalized normal distribution proves highly effective in encapsulating uncertainties arising from shifts due to noise and changes in modality, which often manifest as outliers within datasets. The loss in Eq.\ref{eq:6} consists of a fidelity term along with general constraints on the shape and scale of the residual distribution to prevent divergence to infinity. But, given that I2I translation can be characterized by a lack of a unique stable solution, incorporating explicit constraints on the parameters of the residual distribution into the objective function, typically in the form of a penalty benefits to progressively refine uncertainty estimation. This is discussed in the next section.

\subsection{Uncertainty-Aware Regularization}
\begin{figure}[!htbp]
    \centering
    \includegraphics[width=4cm, height=3.5cm]{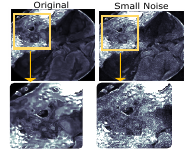}
    \caption{Uncertainty estimate is sensitive to small changes in input and network parameters, resulting in potentially noisy maps.}
    \label{fig:noisy_pixels}
\end{figure}

We operate under the benign assumption that for good reconstructions pixel-residuals exhibit piece-wise continuity similar to images, implying that since adjacent pixels within one image region show minimal discrepancies their residuals should also be similar, unless influenced by noise. Therefore, large residuals can come from pixels of two types, the pixels that the network actually finds hard to reconstruct to due to lack of knowledge or data drifts, and, the spurious pixels that might not strictly correspond to difficulty in reconstruction but end up having high values. To illustrate this better, we simulate this effect by injecting a small amount of noise in an image (Fig. \ref{fig:noisy_pixels}), such that the corruption is visually imperceptible in the image and predict the uncertainty. As expected, the aleatoric uncertainty map is adversely affected, even with comparable reconstructions. Although sensitivity to noise in input data is generally advantageous, excessive sensitivity within the anticipated noise spectrum can result in unreliable and inaccurate uncertainty predictions. We propose to suppress this spurious component for a more accurate estimation of uncertainty by penalizing large differences in the predicted residual distributions for neighboring pixels.

This prior assumption can be incorporated by adding a penalty/regularization term that discourages significant deviations between the predicted residual distributions of adjacent pixels, inline with the expectation that neighboring pixels in an image are likely to have similar residuals, unless there is noise or an edge. Further, while enforcing the above constraint, it is crucial to prevent accidentally suppressing those deviations that occur as a result of the network's incapacity/lack of knowledge to reconstruct an input. This is the interesting case when the network is unsure how to reconstruct the output for one or more regions. Thus, we propose to impose a total-variation based penalty on the estimated shape parameter $\hat{\beta}$ during the learning process, to smooth out noise in the estimated parameters across neighboring pixels while preserving regions of true uncertainty.

For a predicted $\hat{\beta}_i$ image corresponding to input $x_i^b$, the total variation is shown in Eq. \ref{eq:13}.
\begin{equation}
    \label{eq:13}
    TV(\beta_i) = \int_{v} | \nabla\hat{\beta}_i(v)|  dv
\end{equation}
The approximation in two dimension yields,
\begin{equation}
    \label{eq:7}
    R_{\beta_i} = \sum_{jk} \sqrt{(\hat{\beta}_{ij+1 k} - \hat{\beta}_{ijk})^2 + (\hat{\beta}_{ij k+1} - \hat{\beta}_{ijk})^2}
\end{equation}
To prevent derivative singularity, a small positive constant $\epsilon = 10^{-7}$ is added, introducing some blurring, 
\begin{equation}
    \label{eq:77}
    R_{\beta_i} = \sum_{jk} \sqrt{ \epsilon^2 + (\hat{\beta}_{ij+1 k} - \hat{\beta}_{ijk})^2 + (\hat{\beta}_{ij k+1} - \hat{\beta}_{ijk})^2}
\end{equation}
\begin{figure*}
   \centering
    \includegraphics[width=12cm,height=6cm]{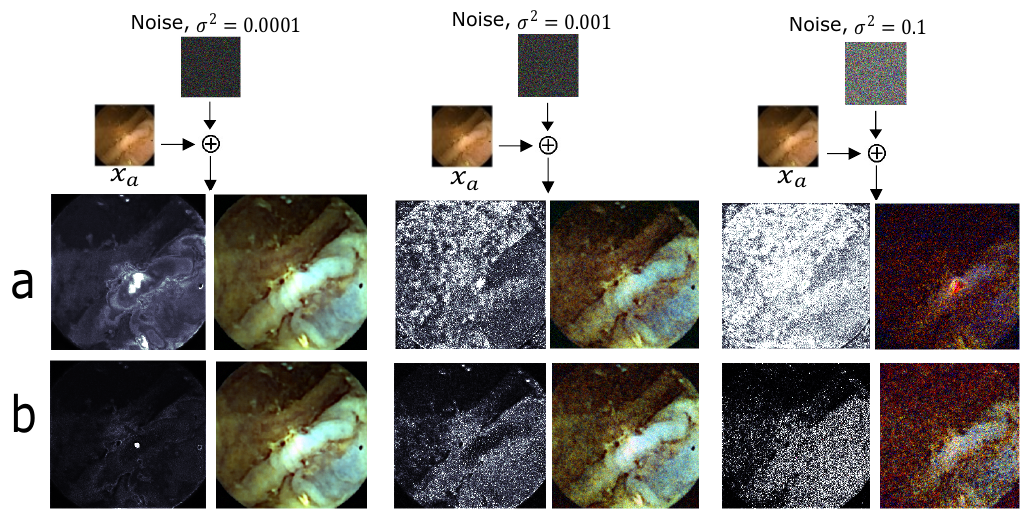}
    \caption{Impact of noise on uncertainty prediction and image reconstruction. The figure illustrates the impact of varying levels of Gaussian noise on image reconstruction and predicted aleatoric uncertainty. Row (a) depicts the non-regularized version, whereas row (b) shows the regularized variant. In the presence of noise, the non-regularized version is more significantly affected, with uncertainty maps rapidly diverging from the original regions of uncertainty. In contrast, the regularized version demonstrates greater robustness to noise.}
    \label{fig:noise_levels}
\end{figure*}
Fig.\ref{fig:regularization_effect} shows the impact of Eq.\ref{eq:77} on uncertainty-estimation over the course of training, highlighting in contrast to its absence. Without regularization, as expected although the parameters $\alpha$ and $\beta$ are predicted, they are not subsequently optimized, even as the image reconstruction continues to improve (due to pixel-wise and GAN loss terms in the objective, equations \ref{eq:8}-\ref{eq:11}). On the other hand, in the regularized variant, residual errors continue to reduce the in tandem with optimizing for parameters $\alpha$ and $\beta$, resulting in cleaner and visually more interpretable uncertainty maps. One notable effect of the UAR penalty on the map is the accentuation of edges around uncertain structures, owing to the edge-preserving nature of total-variation. This facilitates visual interpretation of uncertainty maps, as they correspond to image structures and their uncertainty levels.

\subsubsection*{Model} 
Our model is composed of a single conditional GAN consisting of a generator $G(\cdot; \theta_G)$, and a discriminator $D(\cdot; \theta_D)$. The discriminator follows the commonly used patch architecture \cite{isola2017image, li2016precomputed} while the generator is based on U-Net \cite{ronneberger2015u}. Like \cite{upadhyay2021uncertainty}, the generator outputs spatially varying $\alpha$ and $\beta$, along with the output image from domain $\mathcal{B}$. In addition to the negative log likelihood loss (Eq. \ref{eq:6}) and chosen variation-based regularization (Eq. \ref{eq:7}), the generator is trained using the adversarial loss $L_{adv}$ defined as a mean-squared error between the discriminators predictions for the generated image (Eq. \ref{eq:8}) against the label vectors of ones. This formulation is an alternative to the commonly used cross-entropy loss.
\begin{equation}
    \label{eq:8}
    L_{adv} = \frac{1}{n} \sum_i \mathrm{MSE}(\hat{x}^b_i, 1) 
\end{equation}
Finally, an additional L1-fidelity term, $  L_1 = |x^b_i - \hat{x}^b_i|$ is added to enforce pixel-level reconstruction fidelity between the image and its reconstruction.
Thus, the total loss for the generator is given by
\begin{equation}
    \label{eq:10}
    L_{G} = w_{L_1} L_1 + w_{adv} L_{adv} + w_{nll} L_{nll} +  \lambda  R_{\beta_i}
\end{equation}
where $w_{L_1}$, $w_{adv}$, $ w_{nll}$ and $\lambda$ are the respective weights for each term.
The discriminator is trained using the above-mentioned mean-squared error, with target vectors one for real images and zeros for the generated images.
\begin{equation}
    \label{eq:11}
    L_D = \frac{1}{2} \left[ \frac{1}{n} \sum_i \mathrm{MSE}(\hat{x}^b_i, 0) + \frac{1}{n} \sum_i \mathrm{MSE}(x^b_i, 1)\right]
\end{equation}


\subsection{Training Details and Evaluation Metrics}
We test UAR on two datasets, a new WCE dataset and a public colonoscopy CPC-paired dataset \cite{ma2022CPC}. From the WCE dataset, 5,000 images were utilized for training, and 5,000 for validation. All results are reported on a test-set of another 5,000 image pairs, which is further divided into three subsets for comprehensive evaluation. The training and validation images are sourced from WCE videos of seven patients, while the test images are obtained from three new patients, potentially containing new or different abnormalities. The hyperparameters optimized on the WCE dataset were also effective for the CPC-paired dataset. Consequently, the CPC-paired dataset was split into training and testing subsets (80:20), with results reported on the test set.

Both the discriminator and generator utilize the Adam optimizer with an initial learning rate of $10^{-4}$, following a cosine annealing schedule for learning rate adjustment. The outputs of the discriminator are passed through an average pooling layer before applying the MSE loss (equations \ref{eq:8} and \ref{eq:11}). We found that results improved when the variation-based regularization (Eq. \ref{eq:7}) was activated a little later in the initial learning phase, giving the network a chance to predict unregularized values for $\alpha$ and $\beta$. Thus, the total variation regularization is activated around epoch 5. Through experimentation, we found that a value of $10^{-12}$ for the regularization weight, $\lambda$ in Eq. \ref{eq:10} yielded satisfactory results, with room for further optimization and performance improvements (more details in Section \ref{sec:ablation_lambda}). Other weights in Eq.\ref{eq:10} are $w_{L_1} = 1$, $w_{adv} = 10^{-3}$ and $ w_{nll} = 10^{-4}$.
All models were trained with an image size of $490 \times 490$ and a batch size of 4, using twin-titan RTX GPUs with 48~GB of RAM, achieving a processing speed of approximately 20 images per second. The UAR term can be integrated with minimal computational overhead, as it involves only element-wise operations on the grayscale maps of $\beta$ resulting in execution speeds comparable to those of L1 loss.

To assess the quality of the generated images, we report the results on four metrics, namely Peak Signal-to-Noise Ratio (PSNR), Structural Similarity Index (SSIM) \cite{SSIM}, Relative Root Mean Squared Error (RRMSE) and Learned Perceptual Image Patch Similarity (LPIPS) \cite{zhang2018unreasonable}. While, SSIM, PSNR and RRMSE are more common, we use LPIPS additionally as it has shown to correlate better with human visual perception \cite{zhang2018unreasonable} over pixel-wise metrics. 

\subsection{Dataset}
\label{sec:dataset}
In this work, we introduce a new paired image-to-image translation dataset for capsule endoscopy. The dataset facilitates the translation between original WCE images and their corresponding Flexible Spectral Imaging Color Enhancement (FICE) mode images, and vice versa.

The images were collected during capsule endoscopy trials conducted on patients at Innlandet Hospital, Norway. The trials involved the use of a capsule endoscope to capture images of the gastrointestinal tract in established patients, which were later converted to their corresponding FICE versions using a WCE diagnostic software called rapid reader.
The dataset comprises over 15,000 pairs of carefully curated WCE images, which can be employed for paired as well as unpaired image-translation methods. The dataset is available at \url{https://doi.org/10.18710/BSXNA1}.

\section{Results}
\label{sec:results}
This section presents the qualitative and quantitative evaluation of our method on the two datasets. 
The approach is tested on three types of commonly occurring noises: Gaussian, Uniform and Impulse (also called salt and pepper noise) at different levels. The baseline corresponds to I2I-translation without regularization as in \cite{upadhyay2021uncertainty}.

Fig.\ref{fig:noise_levels} shows the qualitative effect of increasing levels of noise on the predicted uncertainty and reconstruction. Comparing the reconstructed image, it is seen that the regularized variant results in a more visually coherent reconstruction even at high noise levels, as compared to the non-regularized method.

\begin{figure}[!htbp]
    \centering
    \includegraphics[height=10cm, width=6.5cm]{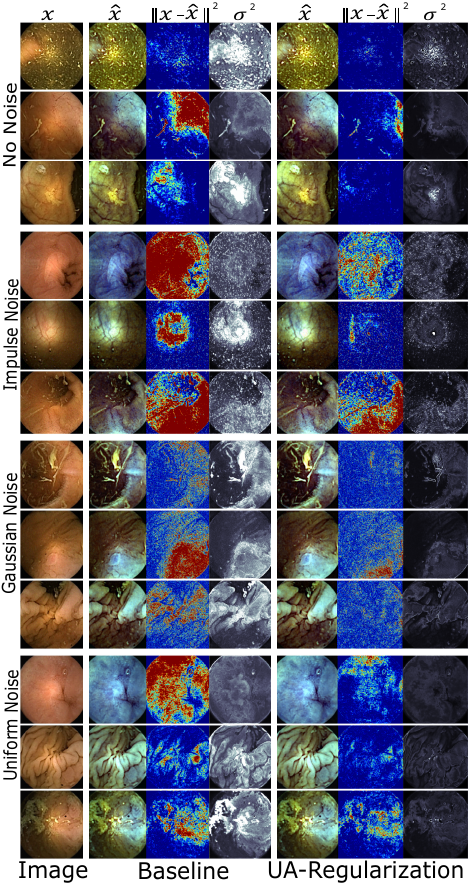}
    \caption{Qualitative Comparison. As can be seen from columns 3 and 6, while the regions of residual errors are consistent between the regularized and non-regularized variant, UAR shows consistently low residual errors. Correspondingly, the uncertainty maps are less-noisy and more structurally coherent.}
    \label{fig:resul}
\end{figure}

Fig.\ref{fig:resul} shows, the residual errors and the uncertainty maps derived from the two methods, under the impact of noise. As seen in columns 4 and 7 ($\sigma^2$), UAR generates less noisy uncertainty maps, consistent with the distinctive features within the images, while reducing the residual errors (columns 3 and 6 ($||x-\hat{x}||^2$)). 

Given the consistent capture modality and similar patient population, the primary source of noise in the data is the added noise itself. If the predicted uncertainties accurately reflect this, they should be low for familiar image structures and higher in noise-affected regions. Uncertainty maps generated using UAR adhere to this expectation. Conversely, in the absence of regularization, the uncertainties are uniformly high, obscuring relative differences in uncertainty and hindering interpretability.
Further, we analyze the quantitative impact of UAR on reconstruction quality in Table \ref{tab:quant} and \ref{tab:cpc}. It is seen that the effect of UAR is overall positive on image reconstruction with equivalent or better SSIM and PSNR values, across different noise types and levels. The regularization also consistently improves the LPIPS and RRMSE metrics, across both datasets.
\begin{table}[htbp]
    \renewcommand*{\arraystretch}{2.1}
    \resizebox{0.489\textwidth}{!}{
        \begin{tabular}{ |c |c | l r r r r| r|}
             \cline{1-7}
             & & \multicolumn{1}{c}{\textbf{Approach}} & \multicolumn{1}{c}{\textbf{SSIM $\uparrow$}}            & \multicolumn{1}{c}{\textbf{PSNR $\uparrow$}}    & \multicolumn{1}{c}{\textbf{LPIPS $\downarrow$}} & \multicolumn{1}{c|}{\textbf{RRMSE $\downarrow$}}     \\ 
             \cline{2-7}
             & \parbox[t]{2mm}{\multirow{2}{*}{-}} &Baseline                 &  0.925 & 28.714  & 0.128   & 0.174 \\ 
          & &UAR (Ours)          & \textbf{0.931} & \textbf{29.34} & \textbf{0.126}  & \textbf{0.148} \\ 
            \cline{2-7}
          \parbox[t]{2mm}{\multirow{2}{*}{\rotatebox[origin=c]{90}{Gaussian}}} & \parbox[t]{2mm}{\multirow{2}{*}{\rotatebox[origin=c]{90}{$\mathcal{N}(0,0.001)$}}}  &Baseline         & 0.639  & 26.610  & 0.275  & 0.212  \\
        &    &UAR (Ours)          & \textbf{0.650} & \textbf{27.04}  & \textbf{0.261}   & \textbf{0.204}  \\ 
            \cline{2-7}
         &    \parbox[t]{2mm}{\multirow{2}{*}{\rotatebox[origin=c]{90}{$\mathcal{N}(0,0.01)$}}} &Baseline  &  \textbf{0.310} & \textbf{22.240 } & 0.452   & 0.399   \\
        &    &UAR (Ours)          &  0.309 & 22.023 & \textbf{0.431}   & \textbf{0.372} \\  
        \cline{1-7}  
        \hline
        \hline
         \parbox[t]{2mm}{\multirow{3}{*}{\rotatebox[origin=c]{90}{Uniform}}} & \parbox[t]{2mm}{\multirow{2}{*}{\rotatebox[origin=c]{90}{$\mathcal{U}(0,0.1)$}}}  &Baseline        &  0.680 &\textbf{ 27.007 }  & 0.287  &  0.212 \\
        &    &UAR (Ours)          & \textbf{0.695}  & 26.421  & \textbf{0.268 }   & \textbf{0.208 }  \\ 
            \cline{2-7}
         &    \parbox[t]{2mm}{\multirow{2}{*}{\rotatebox[origin=c]{90}{$\mathcal{U}(0,0.01)$}}} & Baseline  & 0.912  & 28.14   & 0.135  & 0.173    \\
        &    &UAR (Ours)          &\textbf{ 0.928} & \textbf{29.38}  & \textbf{0.134}   & \textbf{ 0.161}  \\  
                \cline{1-7}  
        \hline
        \hline
          \parbox[t]{2mm}{\multirow{3}{*}{\rotatebox[origin=c]{90}{Impulse}}} & \parbox[t]{2mm}{\multirow{2}{*}{\rotatebox[origin=c]{90}{$\mathcal{I}(0.005)$}}}  &Baseline        & \textbf{0.735 } & 26.63  & 0.380  & 0.226\\
        &    &UAR (Ours)          &  0.724 & \textbf{26.91} &  \textbf{0.371 }& \textbf{0.216 }\\ 
            \cline{2-7}
         &    \parbox[t]{2mm}{\multirow{2}{*}{\rotatebox[origin=c]{90}{$\mathcal{I}(0.01)$}}} &Baseline  &\textbf{ 0.601 } & 25.23   & 0.442  & 0.268     \\
        &    &UAR (Ours)          & 0.5847  & 25.26 & \textbf{0.440 }  &  \textbf{0.256}\\  

                \cline{1-7}  
        \hline
        \hline
        \end{tabular}      
        }
    \caption{Impact of uncertainty-guidance on reconstruction quality. UAR consistently achieves lower LPIPS and RRMSE values across various types and levels of noise, with comparable or superior SSIM and PSNR metrics.}
    \label{tab:quant}
\end{table}
\begin{table}[!htbp]
	\renewcommand*{\arraystretch}{1}
	\resizebox{0.489\textwidth}{!}{
		\tiny
		\begin{tabular}{l l l l l l}
			\hline
			\textbf{Model}                     & \textbf{SSIM$\uparrow$}   & \textbf{PSNR$\uparrow$}  & \textbf{LPIPS$\downarrow$} & \textbf{RRMSE$\downarrow$} \\
			\hline Baseline & 0.891 & 35.358 & 0.410 & 0.291 \\
			   UAR (Ours)   & \textbf{0.925} & \textbf{38.297} & \textbf{0.289}  & \textbf{0.221} \\ 
   \hline
		\end{tabular}
	   }
	\caption{Impact of uncertainty-guidance on reconstruction quality on CPC-dataset \cite{ma2022CPC}. Further results in the supplementary.}
	\label{tab:cpc}
\end{table}
\subsection{Impact of Artifacts}
\label{sec:artefacts}
\begin{figure}[!htbp]
    \centering
     \includegraphics[height=10cm, width=7cm]{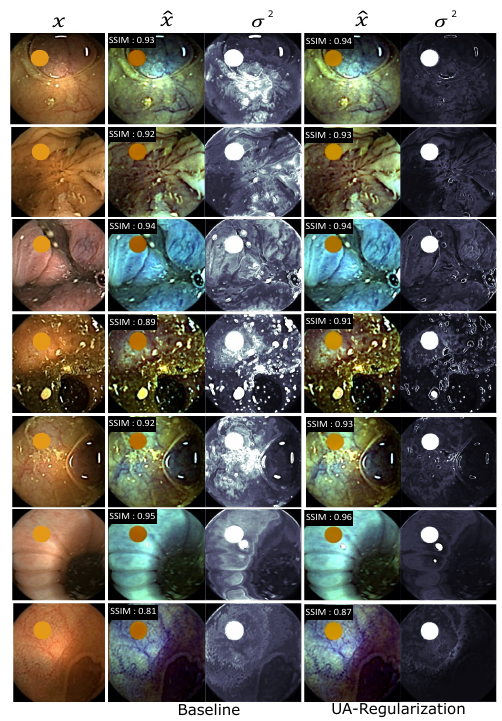}
    \caption{UAR distinctly identifies regions affected by the artificially introduced artifact as having high uncertainty, contrasting with relatively lower uncertainty in unaffected areas. In contrast, the baseline approach shows similar uncertainty levels across different regions, failing to differentiate between previously unseen and unseen image structures.}
    \label{fig:round_artifact}
\end{figure}

Additionally, we evaluate the performance of uncertainty estimation by systematically introducing more pronounced artifacts into the image. Fig.\ref{fig:round_artifact} illustrates images with circular artifacts. The UAR variant prominently displays high uncertainty across the entire artifact region, with comparatively lower uncertainties in other areas of the image. In contrast, the baseline method fails to differentiate the network's confidence between these two regions effectively.

\begin{figure}[!htbp]
    \centering
    \includegraphics[height=10cm, width=7cm]{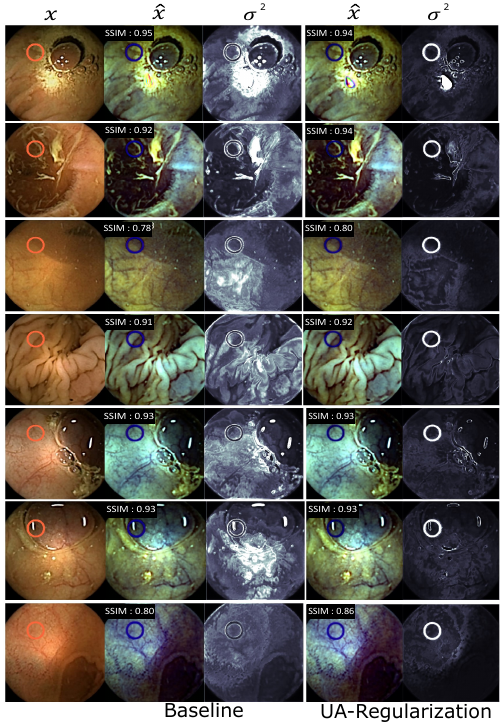}
    \caption{Synthetic artifact and its effect on uncertainty. The uncertainty estimation provided by UAR faithfully reflects the presence of the injected artifact. In the last row, where the artifact is subtle, it induces low uncertainty except around edges in the Baseline method. However, UAR accurately identifies it as an unseen region with high uncertainty.}
    \label{fig:ring_artefact}
\end{figure}

Fig.\ref{fig:ring_artefact} replaces the circular artifact with a ring artifact to examine behaviors near the artifact boundaries. Here again, the baseline method significantly underestimates the uncertainty associated with the artifact, whereas UAR accurately delineates uncertainty regions with precise boundaries (notice last row in Fig.\ref{fig:ring_artefact}).

\section{Ablation}
\label{sec:ablation}

\textbf{Ablation I: Other variation-based losses.}
\label{sec:ablation_losses}
Given that the primary goal of the regularization term is to attenuate spurious variances between nearby pixels, other types of variation-based losses are also conceivable. We experiment with two other variations, and analyze their effect on the uncertainty estimation. We hypothesize that, at a minimum, imposing similar penalties should not negatively affect the reconstruction quality for more faithful uncertainty maps. 

One such penalty could be simply to penalize the squared L2-norm of gradients of the $\beta$ map. This modifies Eq.\ref{eq:77} so that it is differentiable and avoids singularity. However, it comes at the cost of reduced invariance to sharp features, in other words it introduces slight smoothing in the uncertainty map. This variant is referred to as $\text{UAR}_{L2}$.
\begin{equation}
    \label{eq:14}
    R_{\beta_i} = \sum_{jk} \biggl( (\hat{\beta}_{ij+1 k} - \hat{\beta}_{ijk})^2 + (\hat{\beta}_{ij k+1} - \hat{\beta}_{ijk})^2 \biggl)
\end{equation}
Next, we test the regularization of the anisotropic variant of total variation. This is the L1-norm on the gradients of $\beta_i$ given in its discrete form by, 
\begin{equation}
    \label{eq:15}
    R_{\beta_i} = \sum_{jk} \biggl( \sqrt{(\hat{\beta}_{ij+1 k} - \hat{\beta}_{ijk})^2} +   \sqrt{(\hat{\beta}_{ij k+1} - \hat{\beta}_{ijk})^2} \biggl)
\end{equation}
We compare the effects of these different penalty formulations on the reconstruction quality (Table \ref{tab:ablate1}) (as well as qualitatively on the generated uncertainty maps in supplementary). 
Imposing these constraints does not negatively impact the reconstruction quality, as seen in Table \ref{tab:ablate1}. As expected, The uncertainty maps for  $\text{UAR}_\text{L2}$ are smoother compared to UAR and $\text{UAR}_\text{Aniso}$. This is because, while the TV variant prioritizes preserving edges around the uncertain structures, the edges in $\text{UAR}_\text{L2}$  have been smoothed out, though the regions of uncertainty remain consistent. The choice of the best variant may depend on the application's demands or the user preference.

Overall, each of the regularization results in significantly less noisy maps than those without regularization. Given that the testing dataset is similar to the training dataset, low uncertainties are expected, except in the presence of unseen artifacts, as shown in Figures \ref{fig:round_artifact} and \ref{fig:ring_artefact}. For qualitative comparison, please refer to the supplementary material.

\begin{table}[!htbp]
	\renewcommand*{\arraystretch}{1}
	\resizebox{0.489\textwidth}{!}{
		\tiny
		\begin{tabular}{l l l l l l}
			\hline
			\textbf{Model}                     & \textbf{SSIM$\uparrow$}   & \textbf{PSNR$\uparrow$}  & \textbf{LPIPS$\downarrow$} & \textbf{RRMSE$\downarrow$} \\
			\hline Baseline & 0.925 & 28.714 & 0.128 & 0.174 \\
			 $\text{UAR}_{\text{L2}}$  &  0.922 & 27.283 & \textbf{0.126}  & 0.149                 \\
              $\text{UAR}_{\text{Aniso}}$  &  0.927 & \textbf{29.825} & 0.133 & 0.215                 \\
			 UAR    & \textbf{0.931} & 29.340 & \textbf{0.126}  & \textbf{0.148} \\ 
   \hline
		\end{tabular}
	   }
	\caption{Effect of different variation-based penalties on WCE data.}
	\label{tab:ablate1}
\end{table}

\textbf{Ablation II :  $\lambda$.}
\label{sec:ablation_lambda}
We conducted experiments with three values, $10^{-12}, 10^{-7}$ and $10^{-4}$, to capture behaviors across a wide range. For a high value of $\lambda = 10^{-4}$, the regularization effect on $\beta$ is excessively strong. This causes the predicted $\beta$ values to become too similar, suppressing any disparities. Conversely, $\lambda = 10^{-7}$ strikes a balance, offering effective regularization without excessively homogenizing the $\beta$ values. In contrast, using $\lambda = 10^{-12}$ as employed in this study reflects a cautious approach, yielding satisfactory results. We anticipate that the optimal value for $\lambda$ to be within the range $[10^{-7}, 10^{-12}]$ (details in supplementary).
\section{Conclusion and Limitations}
In this work, we presented an end-to-end model for I2I translation that integrates an uncertainty-aware regularization.
UAR aims at ensuring that the model's confidence levels are clearly delineated and easily interpretable while improving the overall reconstruction quality, thereby facilitating better decision-making in safety-critical applications. Through systematic evaluation and ablation studies, we demonstrated that our approach maintains high fidelity in familiar regions while accurately identifying and quantifying uncertainty in novel situations. This paper employs a basic conditional GAN for I2I translation, but more advanced architectures and improved reconstruction losses could enhance translation quality. Since UAR is model-agnostic, it can be seamlessly integrated with these improvements. Additionally, we plan to involve more clinical experts to assess the quality of uncertainty maps, complementing the current qualitative and quantitative evaluations in the future.
\newpage
{\small
\bibliographystyle{ieee_fullname}
\bibliography{egbib}
}

\end{document}